%% file: main_arxiv.tex
\documentclass[10pt,twocolumn,letterpaper]{article}

\usepackage[pagenumbers]{iccv} % To force page numbers, e.g. for an arXiv version
\usepackage[accsupp]{axessibility}  % Improves PDF readability for those with disabilities.

\definecolor{iccvblue}{rgb}{0.21,0.49,0.74}
\usepackage[pagebackref,breaklinks,colorlinks,allcolors=iccvblue]{hyperref}
\usepackage{amsmath}
\usepackage{graphicx}
\usepackage{float}
\usepackage{bbold}
\usepackage{subcaption}
\usepackage{multicol}
\usepackage{multirow}
\usepackage{makecell}
\usepackage{booktabs}
\usepackage{amssymb}
\usepackage{colortbl}

\newcommand{\drule}{\specialrule{0.2pt}{1pt}{1pt}
            \specialrule{0.2pt}{0pt}{\belowrulesep}}

\def\paperID{4054} % *** Enter the Paper ID here
\def\confName{ICCV}
\def\confYear{2025}

\title{Personalized OVSS: Understanding Personal Concept \\ in Open-Vocabulary Semantic Segmentation}

\author{
Sunghyun Park$^{\ast}$\; Jungsoo Lee\thanks{These two authors contributed equally to this work.  ${}^\dagger$Project lead. ${}^\ddagger$Qualcomm AI Research is an initiative of Qualcomm Technologies, Inc.\vspace{-0.1cm}} \; Shubhankar Borse \; Munawar Hayat \\ Sungha Choi$^\dagger$\; Kyuwoong Hwang\; Fatih Porikli \vspace{0.05cm}\\
Qualcomm AI Research$^\ddagger$\\
\texttt{\footnotesize\{sunpar, jungsool, sborse, hayat, sunghac, kyuwoong, fporikli\}@qti.qualcomm.com}
}
\begin{document}

\makeatletter
\g@addto@macro\@maketitle{
% \begin{strip}
    \centering
    \vspace{-0.4cm}
    \includegraphics[width=0.99\textwidth]{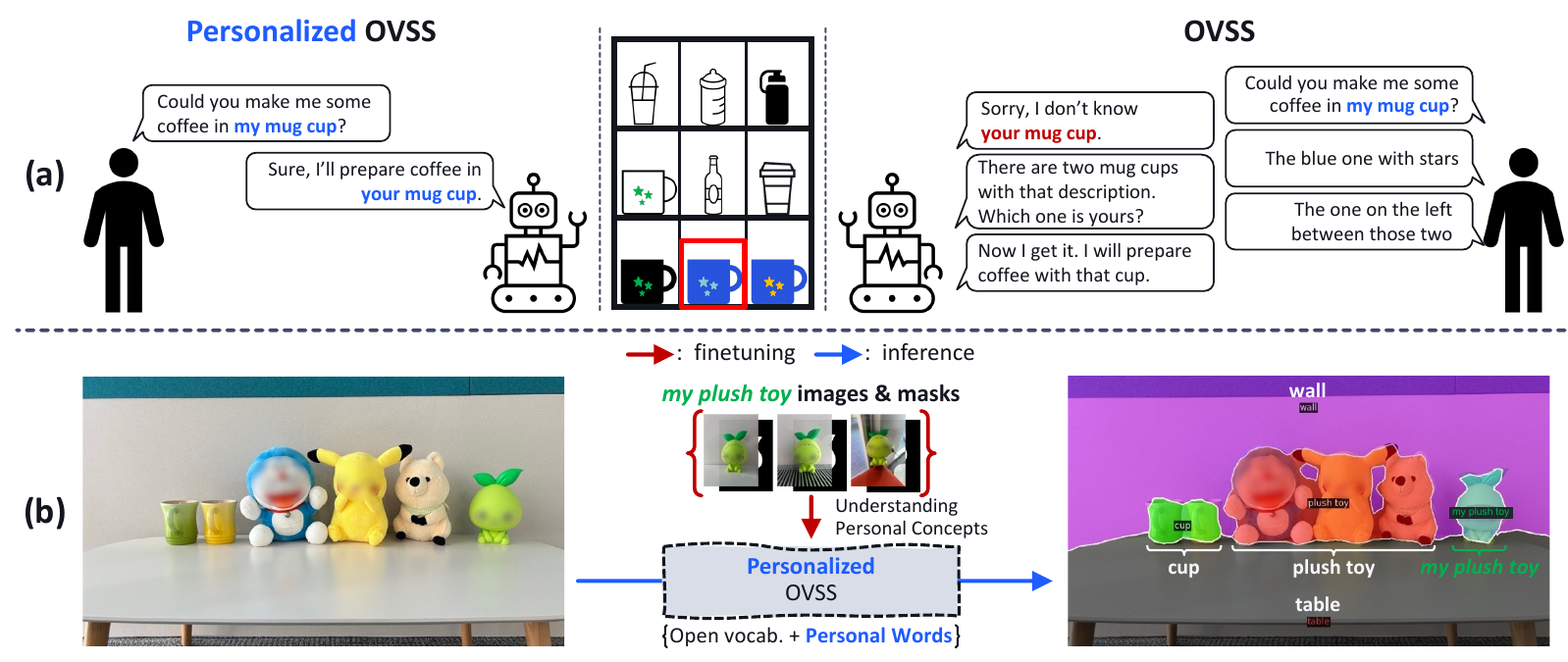}
    \vspace{-0.3cm}
    \captionof{figure}{Description of personalized open-vocabulary semantic segmentation (personalized OVSS). (a) While existing OVSS models require detailed descriptions to find `my mug cup', personalized OVSS enables to find it directly. (b) Given a few images and masks of my personal concept (\eg my plush toy), our proposed framework finetunes an existing OVSS model to distinguish the personal concept among objects within the same class (\eg my plush toy among multiple plush toys).}
    \label{fig:teaser}
    \vspace{0.3cm}
}
\maketitle

\begin{abstract}
    While open-vocabulary semantic segmentation (OVSS) can segment an image into semantic regions based on arbitrarily given text descriptions even for classes unseen during training, it fails to understand personal texts (\eg `\textit{my} mug cup') for segmenting regions of specific interest to users. 
    This paper addresses challenges like recognizing `my mug cup' among `multiple mug cups'.
    To overcome this challenge, we introduce a novel task termed \textit{personalized open-vocabulary semantic segmentation} and propose a text prompt tuning-based plug-in method designed to recognize personal visual concepts using a few pairs of images and masks, while maintaining the performance of the original OVSS.
    Based on the observation that reducing false predictions is essential when applying text prompt tuning to this task, our proposed method employs `negative mask proposal' that captures visual concepts other than the personalized concept.
    We further improve the performance by enriching the representation of text prompts by injecting visual embeddings of the personal concept into them.
    This approach enhances personalized OVSS without compromising the original OVSS performance. 
    We demonstrate the superiority of our method on our newly established benchmarks for this task, including FSS$^\text{per}$, CUB$^\text{per}$, and ADE$^\text{per}$.
\end{abstract}

\vspace{-0.8cm}
\section{Introduction}

\begin{figure*}[t!]
    \centering
    \includegraphics[width=\textwidth]{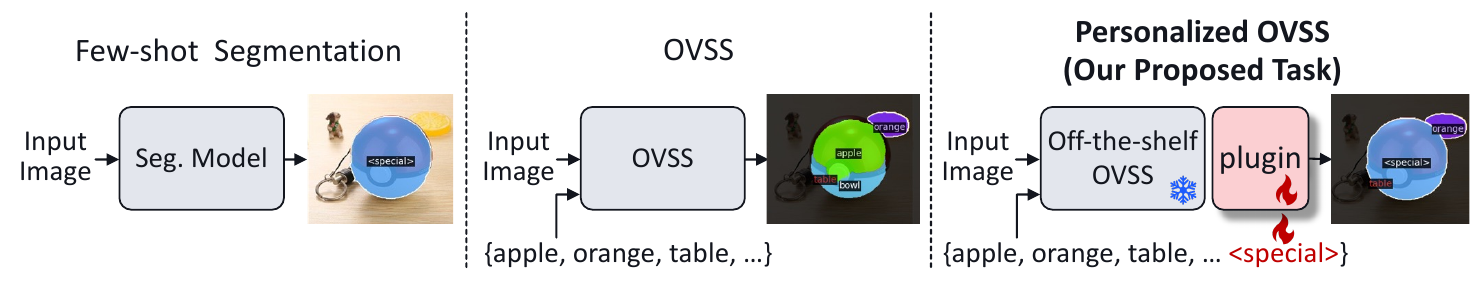}
    \vspace{-0.5cm}
    \caption{How personalized open-vocabulary semantic segmentation (personalized OVSS) differentiates with existing tasks. Few-shot segmentation utilizes a few images and masks for capturing a certain visual concept (\eg pokeball), without considering making predictions with an arbitrary set of words.
    While OVSS predicts objects with an arbitrary set of vocabulary (\eg table and orange), it fails to recognize a given visual concept (\eg pokeball).
    Unlike the two tasks, personalized OVSS recognizes the personal concept along with its original OVSS predictions.
    Additional to the novel task, we also propose a plug-in method applicable to existing off-the-shelf OVSS methods.}
    \vspace{-0.4cm}
    \label{fig:task}
\end{figure*}

% introducing open-vocab segmentation
Due to the recent advances in large-scale vision-language models (\eg CLIP~\cite{clip}), open-vocabulary semantic segmentation (OVSS)~\cite{spnet,odise}, the task of performing segmentation with a set of arbitrary open-vocabulary texts, has recently shown large improvements. 
Unlike traditional semantic segmentation~\cite{chen2018encoder} that is limited to making segmentation predictions within a fixed set of categories, OVSS enables to segment regions with arbitrary classes that are not used during the training phase. 
Such models are crucial for deploying semantic segmentation models in real-world applications (\eg robot assistants) since we may encounter novel objects that were unseen during training. 

% why personalized open-vocab segmentation is needed (few-shot -> personalized / teaser example)
Despite the recent remarkable improvements in OVSS~\cite{spnet,odise,zegformer,zsseg,segclip,maskclip,san}, segmenting the region based on user's interests using unseen categories has been underexplored.
For example, as shown in Fig.~\ref{fig:teaser} (a), recognizing `\textit{my mug cup}' among a number of mug cups in a cupboard is challenging for existing OVSS studies.
The main reason is that they are designed to distinguish between mug cups and other classes (\eg bottles and plastic cups) rather than identifying a certain mug cup among several mug cups.
This feature is important when considering its application in real-world scenarios.
For example, with current OVSS models, we have to provide detailed descriptions to ask our robot assistant to identify `my mug cup', which is cumbersome. 
On the other hand, personalized OVSS enables the robot assistants to identify `my mug cup' even without such detailed descriptions.

Although there is another group of studies that focuses on learning certain visual concepts using few images and masks, referred to as few-shot segmentation~\cite{few_shot_simpler, few_shot_strong_baseline, few_shot_tsf}, these studies fall short in two key aspects for this particular task.
First, they are not designed to perform open-vocabulary segmentation, which is crucial for real-world applications that need to recognize a wide range of objects.
% \js{While Yo'LLaVA resorted to using large language models for understanding open-vocabulary while capturing a personal concept, understanding personal concept along with open-vocabulary has been underexplored without leveraging such large models in semantic segmentation.}
% Open-vocab을 못한다.
Second, they do not consider identifying the personal visual concept among objects sharing the same category (\eg finding `my plush toy' among other plush toys).
As described in Fig.~\ref{fig:task}, we propose a new task termed \textit{personalized open-vocabulary semantic segmentation}, which aims to segment regions of user's interest by using an arbitrary set of words and learning personal visual concepts with few images and masks.
The main goals of this task are to 1) recognize the given personal concept and 2) maintain the original OVSS performance.
While Yo’LLaVA~\cite{yollava} recently focuses on personalized visual question answering using large multimodal models, learning personal concepts along with open-vocabulary segmentation models trained on large-scale datasets has been underexplored, despite its importance in practical applications requiring personalized object recognition.

% Method explanation 
In this paper, along with our newly proposed task, we propose a \textit{simple yet effective} method that can be applied to off-the-shelf pretrained OVSS models.
We first utilize text prompt tuning, which focuses on learning the newly given visual concept using an additional learnable textual embedding.
While such an approach indeed recognizes the personal visual concept at a reasonable level, it is accompanied by an increased number of false positive predictions.
To address such an issue, we propose a `negative mask proposal' that adds a mask to focus on learning the regions other than the personalized concept.
Through the negative mask proposal, we prevent the model from making overconfident predictions during text prompt tuning. 
Additionally, we further improve the performance by injecting the visual embeddings extracted from a pretrained image encoder (\eg CLIP~\cite{clip}) into the textual embedding. 

The contributions of our work are as follows:
\begin{itemize}
    \item To the best of our knowledge, this is the first work to propose the task termed \textit{personalized open-vocabulary semantic segmentation}, which segments images with both an arbitrary set of words and the given personal visual concept included in a few pairs of images and masks.
    \item We propose a \textit{simple yet effective} plug-in method applicable to existing open-vocabulary segmentation models, which performs text prompt tuning with visual information injected and a negative mask proposal.
    \item We achieve state-of-the-art performance on three newly established personalized open-vocabulary segmentation benchmarks, including FSS$^\text{per}$, CUB$^\text{per}$, and ADE$^\text{per}$. 
\end{itemize}

\section{Related Work}
\noindent\textbf{Open-Vocabulary Semantic Segmentation.}
In order to enable segmentation models to segment images with unseen categories, previous open-vocabulary segmentation models~\cite{spnet, joem, cagnet, zs3net} have concentrated their efforts on mapping the visual features to the semantic space.
Recently, the enhanced capacity of cross-modal alignment (\eg CLIP~\cite{clip}) has led to a proliferation of studies utilizing pretrained vision-language models for OVSS~\cite{lseg, zegformer, zsseg, xpm, maskclip, freeseg, san, odise,kNN_CLIP}. 
Side Adapter Network (SAN)~\cite{san} attaches a lightweight side network to a pretrained CLIP, which focuses on predicting mask proposals while leveraging the features of the original CLIP.
Also, based on the observation that the representation space of the Diffusion model~\cite{ldm} is semantically differentiated and localized well, Open-vocabulary DIffusion-based panoptic SEgmentation (ODISE)~\cite{odise} leverages the image encoder of diffusion models for segmenting images into open-vocabulary labels.
Although such models show superb performance on OVSS, they exhibit shortcomings in segmenting regions of user's interest.
To address these challenges, this paper proposes a plug-in method applicable to off-the-shelf OVSS models, enabling the learning of new visual concepts.

\noindent\textbf{Personalization of Semantic Segmentation.}
Performing semantic segmentation for the objects of specific interests included in one or a few exemplary images, referred to as personalization of semantic segmentation or few-shot segmentation, has recently received significant attention~\cite{slime, matcher, segic, persam, fss1000, seggpt, seem}.
This interest is largely attributed to the recent advent of large vision foundation models (\eg SAM~\cite{sam} or DINOv2~\cite{dinov2}), which are pretrained with large-scale datasets.
By utilizing their well-generalized knowledge, performing segmentation on specific visual concepts, even with fine-tuning a small number of parameters, has become feasible~\cite{persam,seggpt,seem}. 
One straightforward approach for the personalization of semantic segmentation is fine-tuning the text prompts. 
Despite their outstanding performances, previous studies in the personalization of semantic segmentation fail to incorporate their task with open-vocabulary segmentation. 
As aforementioned, semantic segmentation models in real-world applications may encounter unseen classes, so personalization of semantic segmentation with novel categories is essential for their applicability.
While several methods~\cite{palavra, seggpt, seem, kNN_CLIP, hummingbird} have shown the capability to segment novel concepts using few-shot samples and masks, they are either not designed for, or limited in, distinguishing personalized concepts among objects within the same class (\textit{e.g.}, identifying `my bird' among `multiple birds').
% While SEEM~\cite{seem} demonstrates interactive segmentation performance using open vocabulary, it is not a plug-in method that could be applied to an off-the-shelf pretrained OVSS model.
To this end, we propose a simple yet effective plug-in method for personalized OVSS along with our newly established benchmark.

\section{Method}

Before diving into our proposed method, we briefly explain the widely used standard prediction process of OVSS models~\cite{san,odise}. 
We have the textual embeddings for open-vocabulary segmentation $\textbf{T}\in\mathbb{R}^{V\times D}$, where $V$ and $D$ refer to the open-vocabulary size and the feature dimension, respectively.
We also have mask embeddings $\textbf{Z}\in\mathbb{R}^{N\times D}$, where $N$ indicates the number of masks. 
We compute the similarity between words and masks $\textbf{S}=\textbf{T}\cdot\textbf{Z}^{\text{T}}$, where $\textbf{S}\in\mathbb{R}^{V\times N}$.
Then, by using the similarity map $\textbf{S}$ and the mask proposals $\textbf{M}\in\mathbb{R}^{\frac{H}{16}\times\frac{W}{16}\times N}$, we obtain the final prediction output $\textbf{P}=\textbf{M}\times\textbf{S}^{\text{T}}$, where $\textbf{P}\in\mathbb{R}^{\frac{H}{16}\times\frac{W}{16}\times V}$. 
We resize $\textbf{P}$ to the original image size for the evaluation.

\begin{figure}[t!]
    \centering
    \includegraphics[width=0.88\linewidth]{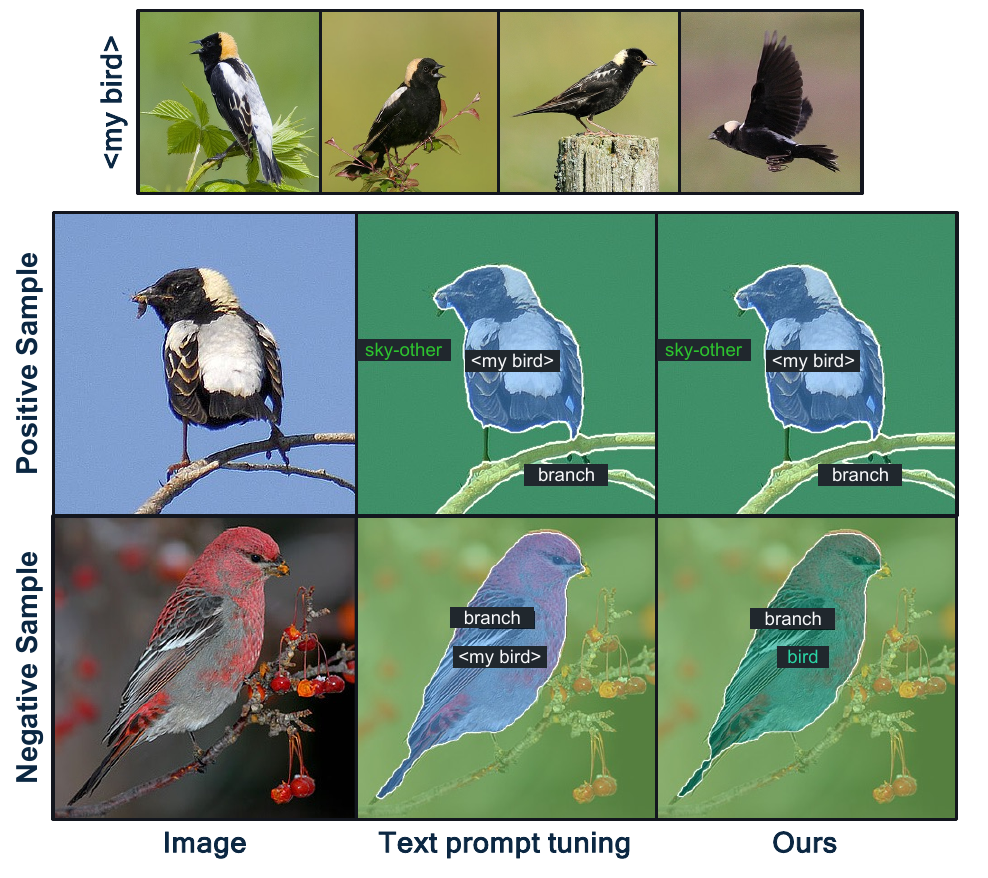}
    \vspace{-0.4cm}
    % \caption{Increased false positives with text prompt tuning. `my bird' and light-blue colored pixels denote the predictions on personal concept. During personalization on `my bird', the first and second row indicates the results on positive and negative samples, respectively. In the second row, we observe increased false positives on the test sample of `$<$my bird$>$' with prompt tuning, where a different bird is also predicted as `$<$my bird$>$' instead of `bird'. In the meanwhile, our method recognizes other birds as `bird' instead of `my bird'.}
    \caption{Increased false positives with text prompt tuning. Light-blue pixels and the label `$<$my bird$>$' indicate predictions on the personalized concept. The first and second rows show results on positive and negative samples, respectively. In the second row, prompt tuning causes false positives, misclassifying a different bird as `$<$my bird$>$' instead of bird. In contrast, our method correctly labels other birds as bird.}
    \vspace{-0.6cm}
    \label{fig:motivation}
\end{figure}

\begin{figure*}[t]
    \centering
    \includegraphics[width=0.9\textwidth]{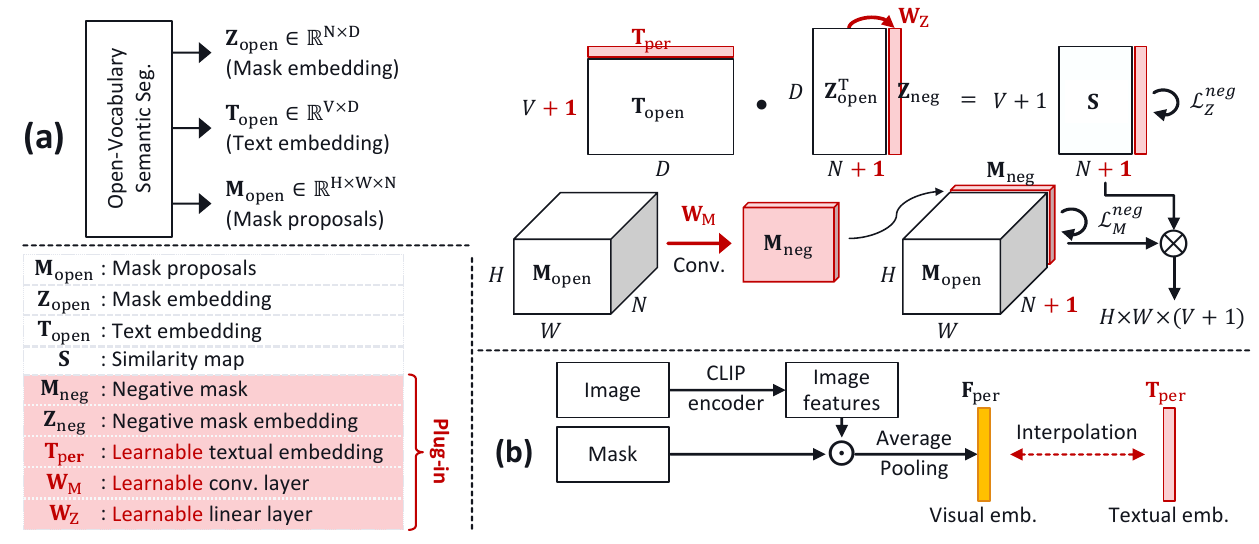}
    \vspace{-0.3cm}
    \caption{Overview of our proposed method. (a) We additionally use a learnable textual embedding $\textbf{T}_{\text{per}}$ to existing textual embedding $\textbf{T}_{\text{open}}$, which focuses on learning the personal visual concept. Given existing mask embedding $\textbf{Z}_{\text{open}}$ and mask proposal $\textbf{M}_{\text{open}}$, we obtain $\textbf{Z}_{\text{neg}}$ and $\textbf{M}_{\text{neg}}$ by forwarding to $\textbf{W}_{\text{Z}}$ and $\textbf{W}_{\text{M}}$, which we add to the original mask embedding and mask proposals, respectively. (b) We obtain a masked visual embedding using an image encoder (\eg CLIP), and we interpolate it with $\textbf{T}_{\text{per}}$ to enrich the information of personal concept.}
    \vspace{-0.4cm}
    \label{fig:method}
\end{figure*}

\vspace{-0.1cm}
\subsection{Text Prompt Tuning}
\vspace{-0.1cm}
\label{method:text_prompt_tuning}
Fig.~\ref{fig:method} describes the overall procedure of our proposed method.
Our method first utilizes text prompt tuning to learn the personal visual concept with a few pairs of images and masks.
Performing text prompt tuning for fine-tuning the pre-trained vision-language models such as CLIP~\cite{clip} has been widely used in various multimodal studies~\cite{coop, cocoop,VoP,learn_to_learn_prompt,TCP,maple}.
While there are various ways for text prompt tuning, we initialize a learnable embedding vector with textual embedding using the class name of the visual concept we want to personalize. 
Then, we forward it to the off-the-shelf OVSS model (\eg SAN~\cite{san}) along with the original textual embeddings for open-vocabulary segmentation $\textbf{T}_{\text{open}}\in\mathbb{R}^{\text{V}\times\text{D}}$. 
In other words, our learnable textual embedding $\textbf{T}_{\text{per}}\in\mathbb{R}^{1\times\text{D}}$ is concatenated to $\textbf{T}_{\text{open}}$, leading us to have final textual embeddings $\textbf{T}=[\textbf{T}_{\text{open}};\textbf{T}_{\text{per}}]$.
For training the learnable textual embedding, we follow the loss functions used in existing open-vocabulary segmentation models. 
Specifically, we utilize a dice loss $\mathcal{L}_{dice}$, a binary cross-entropy loss $\mathcal{L}_{bce}$ for mask generation, and a cross-entropy loss $\mathcal{L}_{cls}$ for mask recognition. 
The total loss is as follows,
\vspace{-0.3cm}
\begin{equation}
    \mathcal{L}_{seg} = \lambda_1 \mathcal{L}_{dice} + \lambda_2 \mathcal{L}_{bce} + \lambda_3 \mathcal{L}_{cls},
\end{equation}
where $\lambda_1$, $\lambda_2$, and $\lambda_3$ are loss weights. 
We set the loss weights following OVSS models we use~\cite{san, odise}.

\subsection{Negative Mask Proposal}
While text prompt tuning recognizes the personal visual concept well, we observe that it may lead to increased false positives.
Fig.~\ref{fig:motivation} well describes our findings, where we examine whether the visual concept of `my bird' is distinguished among various types of birds via text prompt tuning.
When personalizing the OVSS model on the concept of `my bird', text prompt tuning recognizes it well (Fig.~\ref{fig:motivation}, first row).
However, when an image of a different bird is given as the test sample, the model using text prompt tuning also predicts it as `my bird', indicating that it misrecognizes other similar concepts as the personal visual concept, which leads to increased false positives (Fig.~\ref{fig:motivation}, second row). 
Since one of the goals of this task is to maintain the original OVSS performance, reducing false positives is important.
Based on this observation, we propose a negative mask proposal to prevent the model from making false positive predictions.

While not a segmentation work, similar tendency was also observed in recent work, Yo'LLaVA~\cite{yollava}, which states that performing personalization with LLaVA~\cite{liu2023llava} using only positive samples leads the model to always answer that the interested subject is in the photo, regardless of the question.
While Yo'LLaVA necessitates collecting a non-trivial number of negative samples with the same class to address this issue, our method does not require such a labor-intensive process.

Regarding the mask embeddings, we have the original mask embeddings $\textbf{Z}_{\text{open}}\in\mathbb{R}^{\text{N}\times\text{D}}$.
Then, we obtain a negative mask embedding 
$\textbf{Z}_{\text{neg}}\in\mathbb{R}^{1\times\text{D}}$ by performing linear combination of $\textbf{Z}_{\text{open}}$ using the learnable weight $\textbf{W}_{\text{Z}}\in\mathbb{R}^{1\times N}$, which is formulated as,
\begin{equation}
    \textbf{Z}_{\text{neg}} = \textbf{W}_{\text{Z}}\textbf{Z}_{\text{open}}. 
\end{equation}

We concatenate $\textbf{Z}_{\text{neg}}$ to $\textbf{Z}_{\text{open}}$ that gives us $\textbf{Z}=[\textbf{Z}_{\text{open}};\textbf{Z}_{\text{neg}}]$, where  $\textbf{Z}\in\mathbb{R}^{(\text{N}+1)\times\text{D}}$.

Then, we obtain the similarity map $\textbf{S}=\textbf{T}\cdot\textbf{Z}^{\text{T}}$.
Since the goal of $\textbf{Z}_{\text{neg}}$ is to learn words other than the personal visual concept, we supervise it to learn other words equally, which can be formulated using $\textbf{S}$ as,
\begin{equation}
    \mathcal{L}^{\text{neg}}_{\text{Z}} = -\sum^{V}_{i=1,i\neq k} \frac{1}{V-1} \log S[i,j],
\end{equation}
where $j$ and $k$ indicate the index of negative mask embedding and the index of personal concept, respectively.

Given original mask proposals $\textbf{M}_{\text{open}}\in\mathbb{R}^{\frac{H}{16}\times\frac{W}{16}\times N}$, we obtain a mask $\textbf{M}_{\text{neg}}\in\mathbb{R}^{\frac{H}{16}\times\frac{W}{16}\times 1}$ that learns visual concepts other than the personal concept, referred to as the negative mask. 
With a learnable convolutional layer $\textbf{W}_{\text{M}}$, we obtain a negative mask by using $\textbf{M}_{\text{open}}$, which is formulated as,
\begin{equation}
    \textbf{M}_{\text{neg}} = \textbf{W}_{\text{M}}\textbf{M}_{\text{open}}. 
\end{equation}
With the ground truth mask of personal concept $\textbf{M}_{\text{gt}}$, we use $1-\textbf{M}_{\text{gt}}$ as the ground truth mask for the supervision of $\textbf{M}_{\text{neg}}$ using the binary cross-entropy loss, which can be formulated as,
\begin{equation}
    \mathcal{L}^{\text{neg}}_{\text{M}} = -(1-\textbf{M}_{\text{gt}})\log(\textbf{M}_{\text{neg}}) - \textbf{M}_{\text{gt}}\log(1 - \textbf{M}_{\text{neg}}).
\end{equation}
For the final mask proposal, we use $\textbf{M}=[\textbf{M}_{\text{open}};\textbf{M}_{\text{neg}}]$ $\in\mathbb{R}^{\frac{H}{16}\times\frac{W}{16}\times (N+1)}$.

\subsection{Injection of Visual Embeddings}

Based on the observation that utilizing a single modality (\eg vision or language) for prompt tuning leads to a limited performance gain, recent studies~\cite{maple,DAPT,PromptSRC} incorporate both representation spaces of vision and language models for prompt tuning. 
Inspired by such studies, we further improve the performance of personalized OVSS by injecting the visual information into the learnable textual embeddings, as described in Fig.~\ref{fig:method} (b). 
Specifically, we bring a pretrained image encoder $\textbf{I}_{\text{enc}}$ (\eg CLIP image encoder) and forward a given image $\textbf{X}$.
Then, we extract the feature map $\textbf{F}=\textbf{I}_{\text{enc}}(\textbf{X})$ and select feature embeddings from the feature map that correspond to the personal concepts using the mask interpolated to match the resolution of $\textbf{F}$, denoted as $\textbf{M}^{'}_{\text{gt}}$, and average them.
This masked visual embedding including the personal visual concept, $\textbf{F}_{\text{per}}$, is formulated as, 
\begin{equation}
    \textbf{F}_{\text{per}}=\frac{1}{\sum^{HW}_{j=1} \mathbb{1}(\textbf{M}^{'}_{\text{gt}}=1)} \sum_{i=1}^{HW}\textbf{F}\odot \textbf{M}^{'}_{\text{gt}},
\end{equation}
where $\odot$ indicates the element-wise multiplication. 
When multiple images are given, we average $\textbf{F}_{\text{per}}$ of multiple images.
Finally, during text prompt tuning, we replace $\textbf{T}_\text{per}$ with $\textbf{T}_\text{per}^{vis} = \alpha\cdot\textbf{F}_{\text{per}} + (1-\alpha)\cdot\textbf{T}_\text{per}$, which is the interpolation between the masked visual embedding $\textbf{F}_{\text{per}}$ and the personalized textual embedding $\textbf{T}_\text{per}$.
% In addition to the proposed loss functions, we use the original segmentation loss function $\mathcal{L}_{seg}$ used in the off-the-shelf segmentation model we bring.
To this end, our total loss function is formulated as, 
\begin{equation}
    \mathcal{L}_{total}= \mathcal{L}_{seg} + \lambda^{\text{neg}}_{\text{Z}}\mathcal{L}^{\text{neg}}_{\text{Z}} + \lambda^{\text{neg}}_{\text{M}}\mathcal{L}^{\text{neg}}_{\text{M}}.
\end{equation}

\input{tables/main-iou}

\section{Experiments}

\subsection{Experimental Settings}

\noindent\textbf{Datasets.}
We conduct experiments using three datasets: 1) FSS-1000~\cite{fss1000}, 2) CUB-200~\cite{cub-200}, and 3) ADE-20K-847~\cite{ade-20k}.
Specifically, we bring pretrained OVSS models and perform the personalization of a certain visual concept on these three datasets. 
The models are pretrained with COCO Stuff~\cite{coco-stuff}, so we assume that classes included in the three datasets used for evaluation are regarded as open-vocabulary words.

We evaluate the models depending on the number of images $K$ used for personalization, which we set to $K = 1, 3, 5$. 
For the test sets, we include equal number of images including the personal concept and those without the personal concept. 
We intentionally include the images without the personal concept since models that are fine-tuned to recognize personal concepts overconfidently predict other concepts as personal concepts as well.
Preventing such behavior is one of the main goals of our work, so we include images without personal concepts in our test sets. 
We select 30 classes for FSS-1000 and ADE-20K-847, while using all 200 classes for CUB-200.
Since the selected classes and evaluation settings are different from the original datasets, we refer to the processed versions of FSS-1000, CUB-200, and ADE-20K as FSS$^{\text{per}}$, CUB$^{\text{per}}$, and ADE$^{\text{per}}$, respectively.
Further details are included in the supplementary.

\noindent\textbf{Evaluation Metrics.}
We mainly utilize two different evaluation metrics using intersection over union (IoU): IoU$^{\text{per}}$ and mIoU.
IoU$^{\text{per}}$ evaluates how precisely a given segmentation model predicts the personal concept.
Specifically, the regions of the personal concept are considered as the positive labels, while the regions of other classes are all negative labels.
For mIoU, in addition to IoU$^{\text{per}}$, we include IoU of other classes and average the values of IoU.
Since ADE-20K-847 originally includes the ground truth labels for open-vocabulary classes, we use those labels for evaluating mIoU in ADE$^{\text{per}}$.
However, FSS-1000 and CUB-200 do not include the ground truth labels for open-vocabulary classes, so we assume the predictions of pretrained open-vocabulary segmentation models as the ground truth labels for calculating mIoU for FSS$^{\text{per}}$ and CUB$^{\text{per}}$.
The main goal of this work is to improve IoU$^{\text{per}}$ while maintaining a reasonable level of performance on mIoU.

\noindent\textbf{Implementation Details.}
For the pretrained open-vocabulary segmentation models, we utilize two representative open-vocabulary segmentation models, \textbf{S}ide \textbf{A}dapter \textbf{N}etwork (SAN)~\cite{san} and \textbf{O}pen-vocabulary \textbf{DI}ffusion-based panoptic \textbf{SE}gmentation (ODISE)~\cite{odise}.
For SAN and ODISE, we use ViT-B/16 CLIP model and ViT-L/14 CLIP model, respectively.
Since ODISE utilizes the diffusion model, we use the Stable Diffusion v1.3 model~\cite{ldm} for it.
We train $\textbf{T}_{\text{per}}$, $\textbf{W}_{\text{M}}$, and $\textbf{W}_{\text{Z}}$ for 200 iterations.
We initialize the learnable textual embedding with the textual embedding of the label of the personal concept using the text description `a photo of' before the label.
Further details on our experimental settings are included in the supplementary.

\subsection{Quantitative Results}

For quantitative evaluation, we compare open-vocabulary segmentation models without personalization and those with our method.
Specifically, the models without personalization employ only text prompts for recognizing personal visual concept, such as `black footed albatross'.
Table~\ref{tab:IoU} compares IoU$^{\text{per}}$ and mIoU in personalized OVSS.
Across three datasets, our method improves IoU$^{\text{per}}$ when applied to both SAN and ODISE, demonstrating its scalability.
For example, by average, we improve IoU$^{\text{per}}$ of SAN by 12.48, 8.65, 15.79 for FSS$^{\text{per}}$, CUB$^{\text{per}}$, and ADE$^{\text{per}}$, respectively. 
While our method requires a few pairs of images and masks for personal concept, we still improve IoU$^{\text{per}}$ even with one sample, verifying the practicality in real-world applications.

Along with improving IoU$^{\text{per}}$, maintaining the original OVSS performance is also important for personalized OVSS.
Table~\ref{tab:IoU} also shows that our method maintains mIoU for both SAN~\cite{san} and ODISE~\cite{odise} across three datasets. 
This indicates that our method does not compromise the OVSS performance for recognizing personal visual concepts. 
Maintaining the original OVSS performance is crucial for deploying such models in real-world applications since users do not want their models to misrecognize objects that the models used to recognize well.
To the best of our knowledge, this is the first work to establish the experimental setting of personalized OVSS task.
We believe that establishing such an experimental setting itself serves as a large contribution to this field.
Additionally, since this is a newly proposed task, we lack baseline methods beyond the open-vocabulary segmentation models without personalization. 
For further demonstrating the superiority of our proposed method, we compare our method with an approach that replaces the learnable text prompts with visual embeddings, which we report in our supplementary.

\begin{figure*}[t!]
    \centering
    \includegraphics[width=0.8\linewidth]{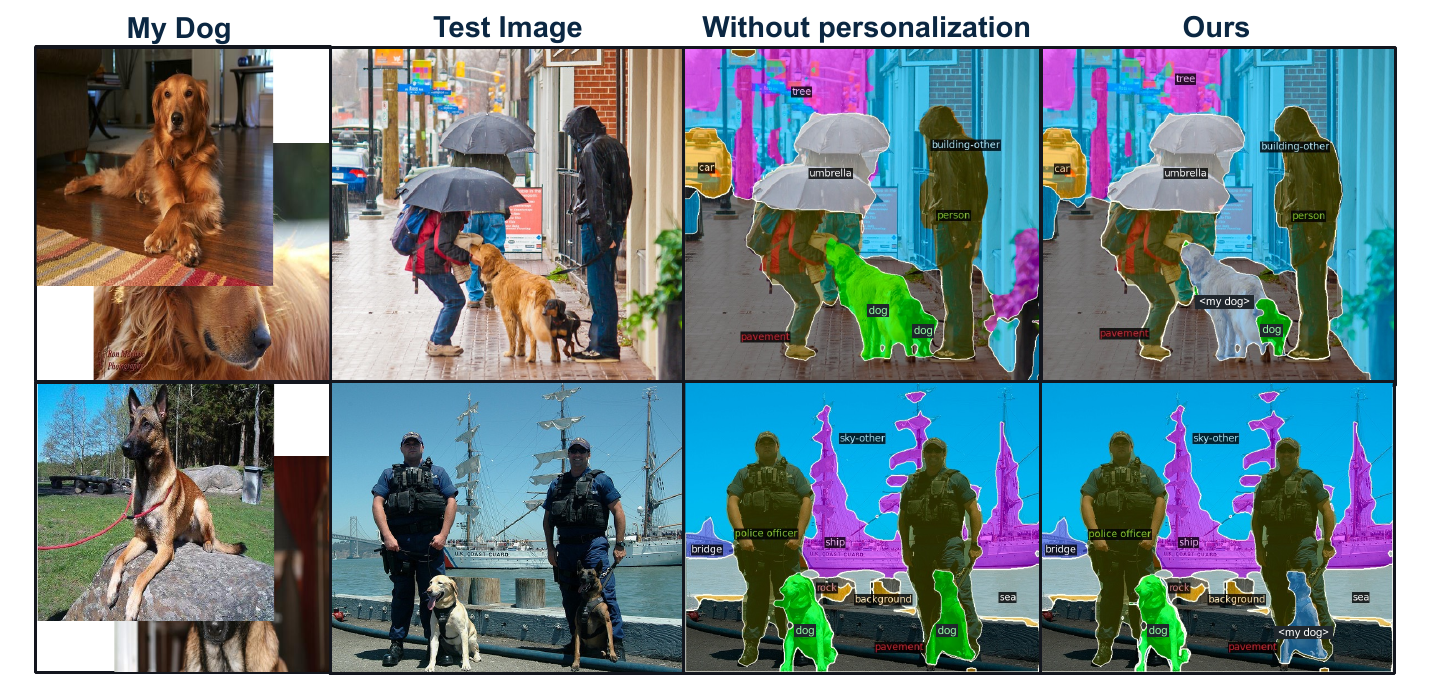}
    \vspace{-0.53cm}
    \caption{Qualitative results of personalized OVSS on test samples from COCO dataset. While existing OVSS models simply recognize different dogs as a dog, personalized OVSS identifies `my dog' among two dogs in the same image.}
    \vspace{-0.4cm}
    \label{fig:coco}
\end{figure*}

\begin{figure*}[t!]
    \centering
    \includegraphics[width=0.82\linewidth]{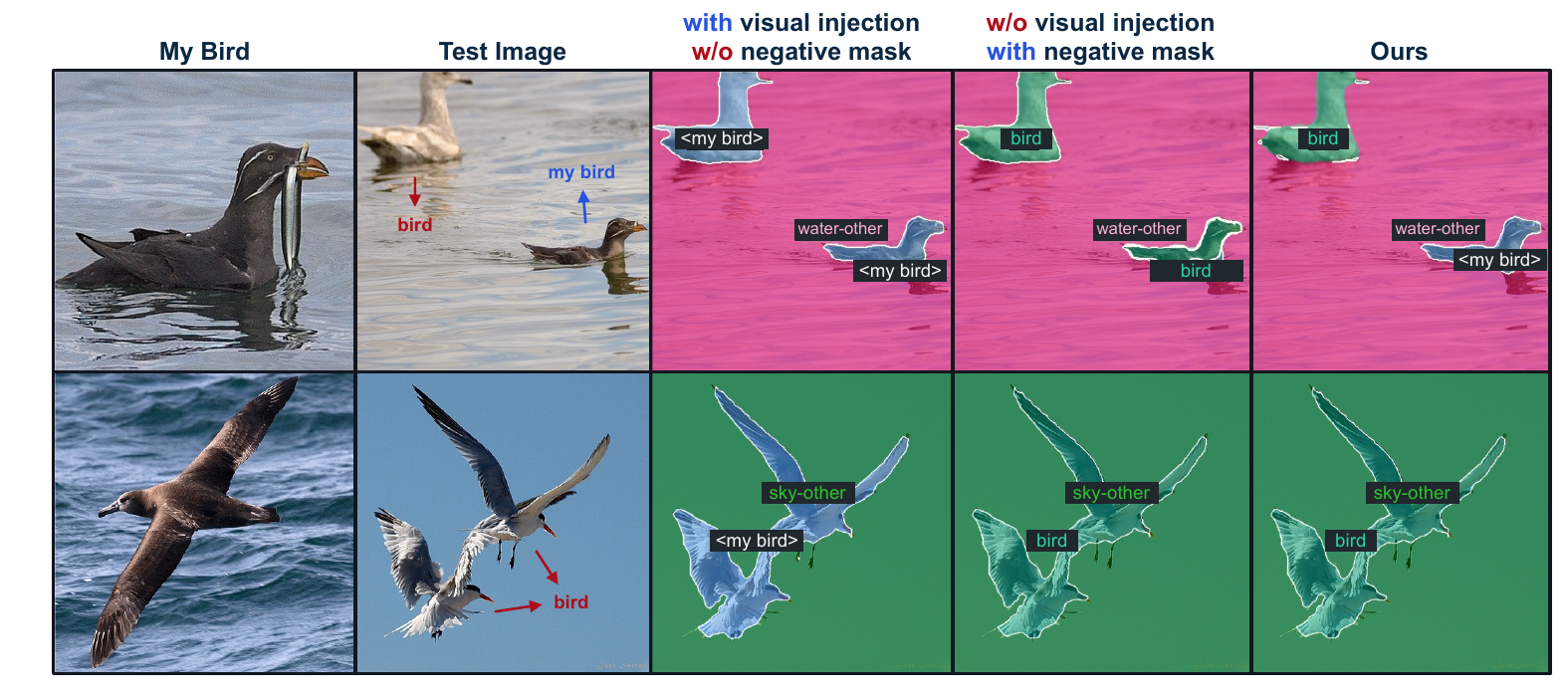}
    \vspace{-0.5cm}
    % \caption{Segmentation results on CUB-200 dataset. We show the effect of each module in our proposed method. While injecting visual embedding to text prompt tuning without negative mask proposal identifies my bird, it also recognizes other birds as my bird. While only utilizing negative mask reduces such false positives, it may fail to recognize my bird without visual embedding injection. Therefore, utilizing both modules is essential.}
    \caption{Segmentation results on CUB-200. We show the effect of each module in our method. Visual embedding injection alone identifies my bird well but increases false positives. Negative mask alone reduces false positives but may miss my bird. Therefore, combining both is essential for accurate segmentation.}
    \vspace{-0.6cm}
    \label{fig:cub}
\end{figure*}

\section{Further Analysis}

\input{tables/main-ablation}

\noindent\textbf{Ablation Study}
For an in-depth analysis of our proposed method, we conduct experiments to investigate how each component of our proposed method contributes to the performance gain.
Specifically, we show how applying text prompt tuning, utilizing negative mask proposal, and injecting visual embedding improves the performance. 
For ablation study, we use CUB$^{\text{per}}$ with $K=5$ and SAN for the dataset and the pretrained model, respectively. 
In addition to mIoU and IoU$^{\text{per}}$, we report IoU$^{\text{per}}_{\text{precision}}$ and IoU$^{\text{per}}_{\text{recall}}$ to analyze how IoU$^{\text{per}}$ improves regarding true and false positive predictions.

% regarding the predictions of true positives and false positives. 
As shown in Table~\ref{tab:ablation}, we observe that applying text prompt tuning improves recognizing the personal visual concept (IoU$^{\text{per}}$ of 69.70) compared to the one without personalization (IoU$^{\text{per}}$ of 68.25).
While text prompt tuning improves IoU$^{\text{per}}$ by increasing IoU$^{\text{per}}_{\text{recall}}$, we observe a large performance degradation on IoU$^{\text{per}}_{\text{precision}}$, indicating significantly increased false positives since the model overconfidently predicts other objects as the personal visual concept.
To address this issue, we utilize the negative mask proposal that improves IoU$^{\text{per}}_{\text{precision}}$ from 74.75 to 80.07, improving IoU$^{\text{per}}$ to 73.71.
On the other hand, while combining text prompt tuning with visual injection improves IoU$^{\text{per}}_{\text{recall}}$, it rather degrades IoU$^{\text{per}}_{\text{precision}}$.
% When we additionally apply the injection of visual embedding, which is our final method, 
Our final method achieves the best performance (IoU$^{\text{per}}$ of 76.80) with reasonable level of both IoU$^{\text{per}}_{\text{precision}}$ and IoU$^{\text{per}}_{\text{recall}}$.

\noindent\textbf{Segmentation Results}
Fig.~\ref{fig:coco} and Fig.~\ref{fig:cub} shows the segmentation results of our method on several samples from COCO dataset~\cite{coco-stuff} and CUB$^\text{per}$, respectively.
% with the effect of each module in our method.
% In the first row, my bird is taken a photo with another bird.
Fig.~\ref{fig:coco} shows that personalized OVSS enables to identify my dog among multiple dogs while OVSS without personalization fails to differentiate my dog with other dogs. 
Fig.~\ref{fig:cub} visualizes the effect of each module in our method.
As shown in the first row of Fig.~\ref{fig:cub}, when performing personalization with only visual embedding injected, the model also captures the other bird as my bird, showing increased false positives.
Additionally, when only the negative mask proposal is applied without the visual embedding injected, it fails to capture my bird.
Therefore, we need both modules to capture my bird while not recognizing other birds as my bird. 
The second row also shows similar tendency when there is a photo without my bird. 
Such an qualitative analysis clearly demonstrates that utilizing negative mask proposal prevents the model from making false positives and visual embedding injection enriches the representation of textual embedding towards the personal concept.

\noindent\textbf{Comparison with Related Approaches}
To further demonstrate that our proposed personalized OVSS is a challenging task, we compare our method with recent 1) open-vocabulary method~\cite{kNN_CLIP} and 2) few-shot segmentation methods~\cite{seem,seggpt}. Tab.~\ref{tab:knn_clip} demonstrates that kNN-CLIP~\cite{kNN_CLIP} using CLIP fails to recognize the personal concept. Also, Tab~\ref{tab:few_shot} compares our method with SEEM~\cite{seem} and SegGPT~\cite{seggpt}, which are recent state-of-the-art few-shot segmentation methods. As shown, while SEEM and SegGPT capture birds well (improved IoU$^{\text{per}}_{\text{recall}}$), they recognize other birds also as the bird we want to personalize (large degradation in IoU$^{\text{per}}_{\text{precision}}$). 
Such results show that even recent models addressing similar tasks struggle to capture my personal concept among objects of same class.

\input{tables/main-knn_clip}
\input{tables/main-few_shot}

\begin{figure}[t!]
    \centering
    \includegraphics[width=0.9\linewidth]{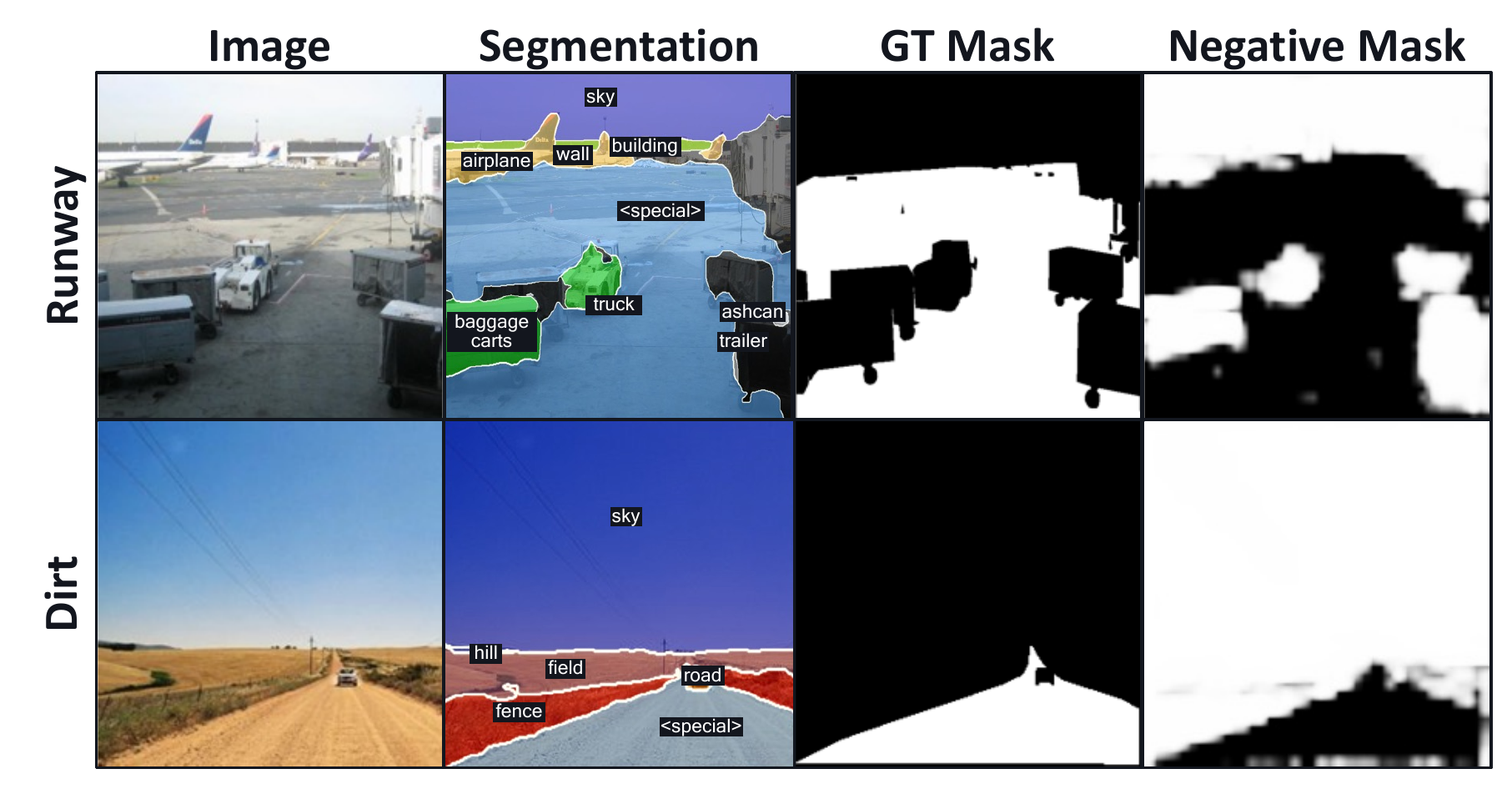}
    \vspace{-0.35cm}
    \caption{Visualization of negative mask proposal. The white and black pixels in the ground truth masks indicate the personal visual concept and other regions, respectively. Our negative mask proposal predicts the opposite, indicating that it learned the concepts other than the personal concept.}
    \vspace{-0.4cm}
    \label{fig:negative_mask}
\end{figure}

\subsection{Visualization of Negative Masks}
We also visualize the negative mask proposals, which we propose in order to prevent the model from making overconfident predictions on other classes when learning the personal visual concept. 
As aforementioned, newly attached negative mask focuses on learning all concepts other than the personal visual concept.
In other words, it outputs low prediction values for regions associated with the personal visual concept and high values elsewhere.
Fig.~\ref{fig:negative_mask} shows that the negative mask well recognizes regions other than the personal visual concept. 
Specifically, white and black pixels in ground truth masks indicate the personal visual concept and other concepts, respectively.
The negative mask proposal outputs the opposite values compared to the ground truth mask, indicating that it outputs low prediction values for the personal visual concept and high values for other concepts.
Adding such a mask focusing on other concepts prevents the model from making false positives, as demonstrated with improved IoU$^{\text{per}}_{\text{precision}}$ in Table~\ref{tab:ablation}.

\subsection{Discussion for Personalized OVSS}

Personalized open-vocabulary semantic segmentation differs from traditional few-shot segmentation~\cite{few_shot_cycle_transformer,few_shot_simpler,few_shot_strong_baseline,few_shot_tsf,seggpt} or class-incremental learning~\cite{class_incremental_essentials, class_incremental_survey_20, class_incremental_survey_24, class_incremental_dual_aug}.
Unlike few-shot segmentation that focuses only on recognizing a specific given concept, personalized open-vocabulary semantic segmentation aims to maintain the capability of open-vocabulary segmentation to segment objects using arbitrary text descriptions while recognizing new personal visual concepts.
To enable various open-vocabulary segmentation models to understand personalized concepts, we design a simple and effective plug-in method based on text prompt tuning.
Furthermore, different from class-incremental learning, which learns new classes sequentially while retaining previous ones without open-vocabulary capability, our method leverages text prompt tuning to capture personal concept based on open-vocabulary capability. 

\begin{figure}[t!]
    \centering
    \includegraphics[width=0.85\linewidth]{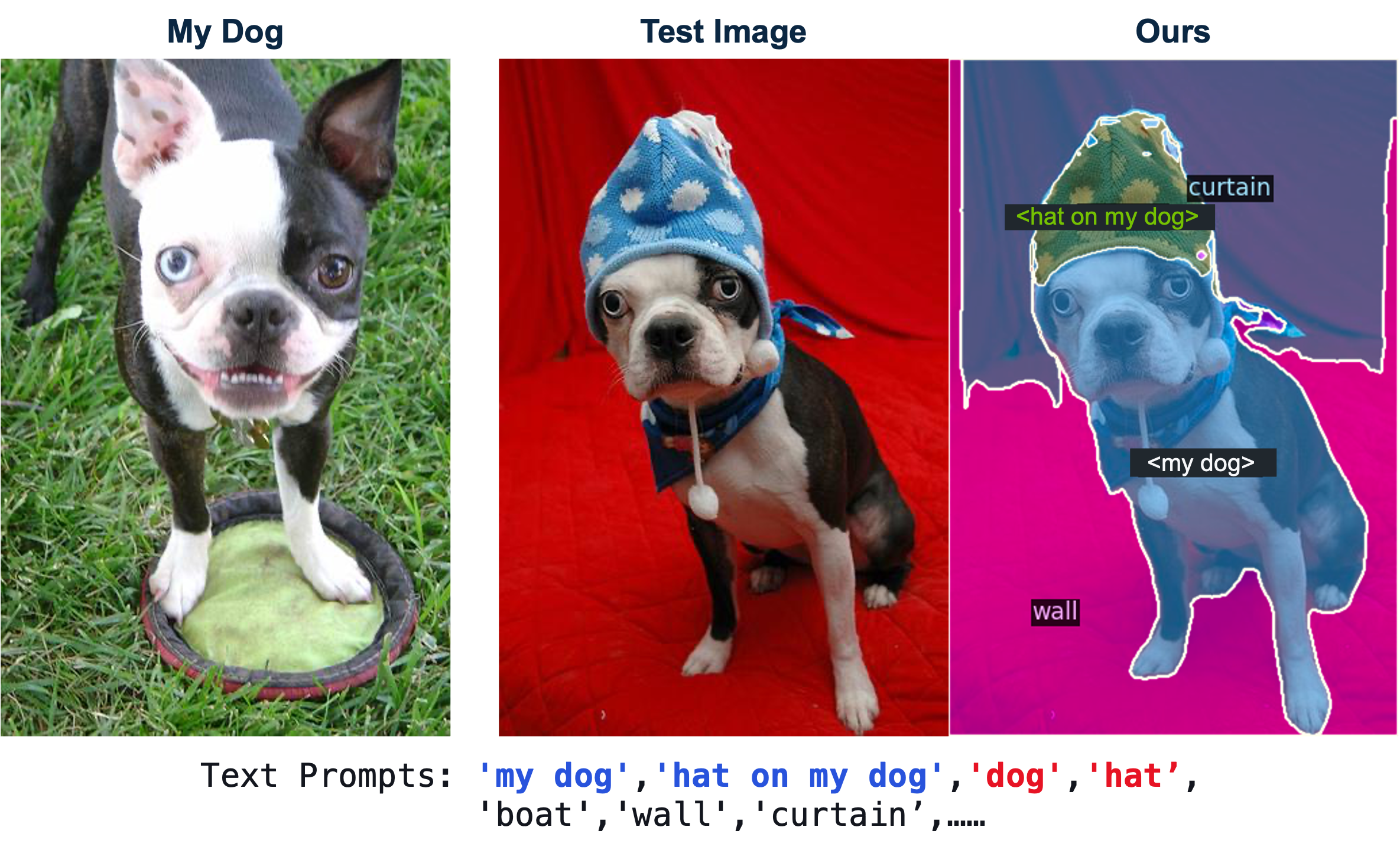}
    \vspace{-0.45cm}
    \caption{Example on recognizing specific object (\ie hat on my dog) by combining the learned personalized prompt (\ie my dog) with the text prompt (\ie hat on). Despite the presence of confusing text prompts such as `dog' and `hat,' it can be observed that the model recognizes the text prompts `my dog' and `hat on my dog' when including the learned personalized prompts.}
    \vspace{-0.5cm}
    \label{fig:dog}
\end{figure}

This approach enables segmentation of specific objects by combining text descriptions with personalized prompts.
For example, our method can recognize both `my dog' and `hat on my dog' instead of `dog' and `hat', as shown in Fig.~\ref{fig:dog}.
We believe that the use cases shown in Fig.~\ref{fig:dog} clearly demonstrates the necessity of our newly proposed task since the current studies in OVSS, few-shot learning, and class-incremental learning fail to recognize `hat on my dog'.
Although we did not include such experiments in our quantitative evaluation due to the absence of the evaluation dataset, we believe that extending our work in such a direction would be an interesting future work.

% Firstly, this task maintains the capability of open-vocabulary segmentation to segment objects using arbitrary text descriptions while recognizing new personal visual concepts.
% \js{While few-shot segmentation or class-incremental learning focus on recognizing a certain given concept, personalized OVSS aims to maintain the original OVSS performance while learning the newly given concept.}
\section{Conclusions}
In conclusion, this study proposes a novel task termed \textit{personalized open-vocabulary semantic segmentation}, which captures the personal visual concept provided from few images and masks utilizing a pretrained OVSS model.
% Unlike existing few-shot segmentation or open-vocabulary semantic segmentation, this task reflects the text descriptions for identifying a personal concept and identifies a certain object among objects sharing the same category.
Unlike existing few-shot segmentation or OVSS, which are limited to segmenting only given visual concepts or fail to understand personalized vocabulary, we extend OVSS models to also comprehend personalized vocabulary, enabling them to identify the personal concept among objects within the same category.
% We demonstrate the importance of this task with an example of asking my robot assistant to find my favorite tumbler or doll among a number of tumblers or dolls.
As aforementioned, such a task is important when asking my robot assistant to find my favorite mug cup among multiple mug cups in a cabinet.
Observing that existing studies do not have an established benchmark for this, we conduct experiments on our newly proposed experimental setting with processed datasets FSS$^{\text{per}}$, CUB$^{\text{per}}$, and ADE$^{\text{per}}$.
This paper also proposes a method including text prompt tuning, negative mask proposal, and injection of visual embedding, which could be applied to existing OVSS methods. % such as SAN~\cite{san} or ODISE~\cite{odise}
% \js{While we do not cover personalizing multiple visual concepts simultaneously, we leave it as one of the future work.}
% While we do not conduct the experiments on personalizing multiple visual concepts simultaneously, we leave it as one of the future work.
Considering the importance of this task, we hope that our work will inspire future researchers to delve into personalized AI systems from recognizing only general words to also understanding personal words, thus moving towards more personalized AI that adapts to individual contexts and preferences.

\newpage

\noindent \textbf{Acknowledgement}
We would like to thank the Qualcomm AI Research team for their valuable discussions.

{
    \small
    \bibliographystyle{ieeenat_fullname}
    \bibliography{reference}
}

\clearpage

\appendix

% Figure - Label Distribution of Datasets
\begin{figure}
\twocolumn[{
\renewcommand\twocolumn[1][]{#1}
    \centering 
    \vspace*{1.1cm}
    \Large{\bf{Supplementary Material}}
    \vspace*{1.3cm}
    
    \includegraphics[width=\textwidth]{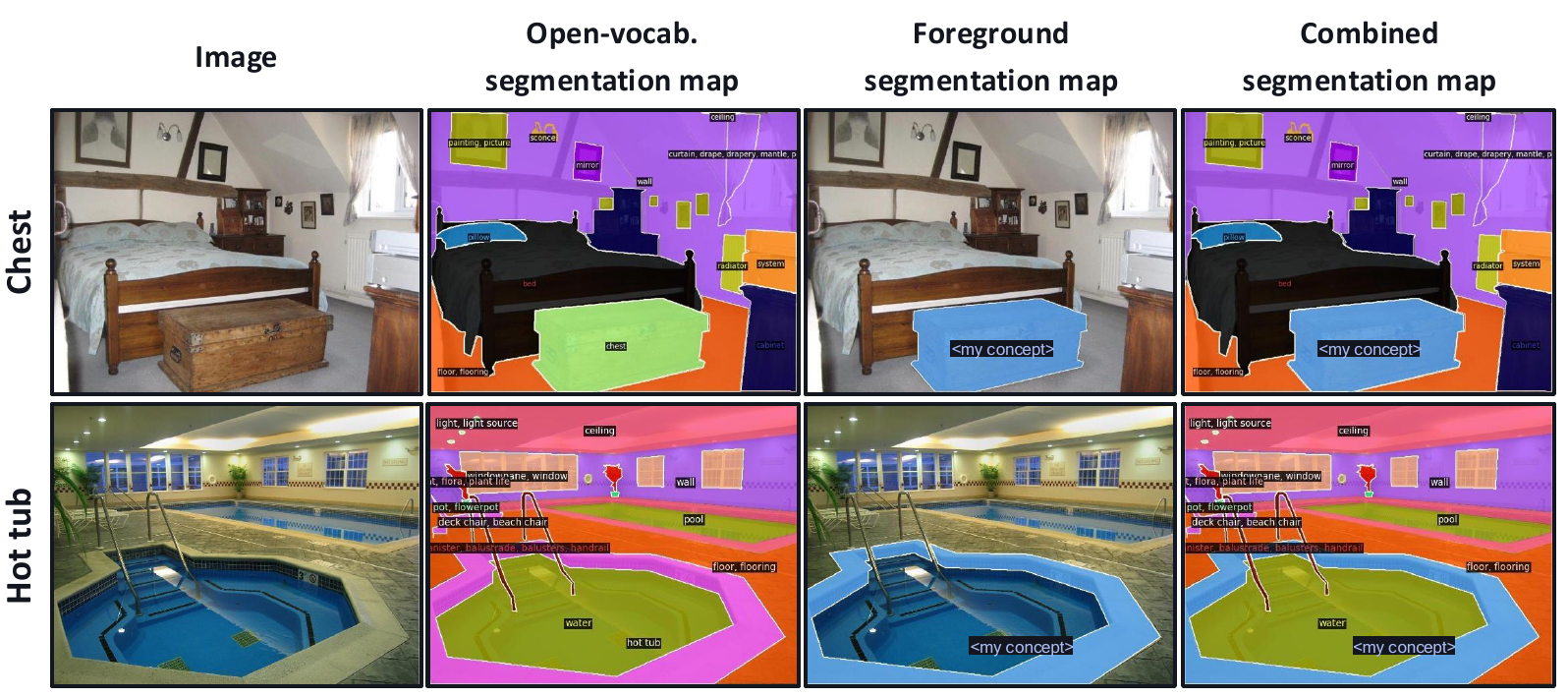}
    \vspace{-0.5cm}
    \captionof{figure}{Examples of segmentation maps we used in ADE$^{\text{per}}$ dataset. We combine the open-vocabulary segmentation map and the foreground segmentation map. 
    Specifically, we annotate the class we want to personalize as the `$<$my concept$>$' category index instead of the original category index (\eg chest and hot tub classes)}
    \label{fig:supple_label}
    \vspace{0.2cm}
}]
\end{figure}

\section{Implementation Details}
\subsection{Further details}
% Implementation Details
In this section, we describe further implementation details of experiments.
For both SAN~\cite{san} and ODISE~\cite{odise}, we use the batch size of 1, 100 number of masks, and $\lambda^{\text{neg}}_{\text{Z}}=0.1$ across all datasets.
Also, regarding the injection of visual embedding, we set the $\alpha=0.1$ for FSS$^{\text{per}}$ and CUB$^{\text{per}}$ and $\alpha=0.01$ for ADE$^{\text{per}}$ for both models.
For SAN, we use the learning rate of 5e-4 and set $\lambda^{\text{neg}}_{\text{M}}=10$ for FSS$^{\text{per}}$ and ADE$^{\text{per}}$ while using $\lambda^{\text{neg}}_{\text{M}}=500$ for CUB$^{\text{per}}$. 
For ODISE,  we use the learning rate of 2e-3 for FSS$^{\text{per}}$ and 1e-4 for CUB$^{\text{per}}$ and ADE$^{\text{per}}$.
We set $\lambda^{\text{neg}}_{\text{M}}=$10.0, 500.0, and 1.0 for FSS$^{\text{per}}$, CUB$^{\text{per}}$, and ADE$^{\text{per}}$, respectively.
We use CLIP~\cite{clip} and Stable Diffusion v1.3~\cite{ldm} for injecting visual embedding for SAN and ODISE, respectively.
The total number of trainable parameters of our method is 0.4M, which denotes that our method requires a negligible amount of additional parameters for personalization.
We conduct all experiments using one A5000 GPU with less than 24GB of GPU memory usage. 
% Seed 얘기할지? 0으로 고정해서 실험하긴 함
% Detailed PyTorch version 등을 서술할지?

\begin{figure*}[t!]
    \centering
    \includegraphics[width=\linewidth]{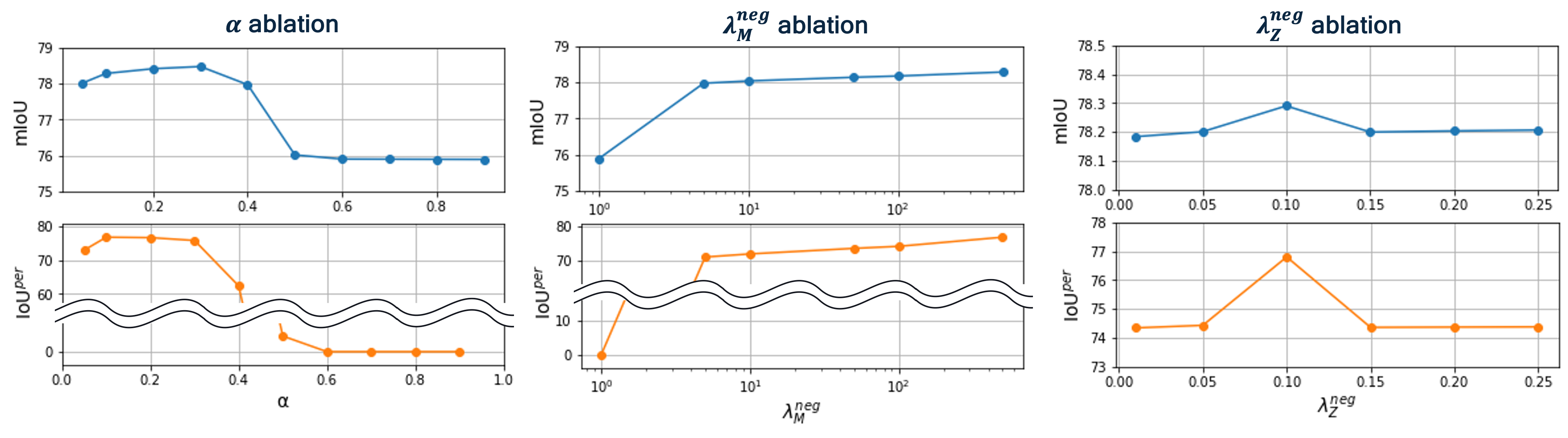}
    \caption{Hyperparameter ablation studies on CUB-200 dataset.}
    % \vspace{-0.4cm}
    \label{fig:hyperparams}
\end{figure*}

\subsection{Dataset Preprocessing}
% Combined Segmentation Maps
Fig.~\ref{fig:supple_label} shows the segmentation maps we utilized.
In order to personalize the open-vocabulary segmentation models, our method only requires the foreground segmentation maps for training.
In other words, during the personalization or training process, it is necessary to provide annotations only for the segments that the user is interested in.
%%%% original %%%% 
% In our experiments, we employ combined segmentation maps to facilitate quantitative evaluation, as shown in Fig.~\ref{fig:supple_label}.
For the quantitative evaluation in our experiments, we employ combined segmentation maps, as shown in Fig.~\ref{fig:supple_label}.
Unlike the previously used open-vocabulary segmentation maps, we annotate categories specified as visual concepts with `<special>' category index for quantitative evaluation.
However, different from ADE$^{\text{per}}$, the FSS$^{\text{per}}$ and CUB$^{\text{per}}$ datasets do not include open-vocabulary segmentation maps and only contain foreground segmentation maps.
%%%% original %%%% 
% Consequently, for the quantitative evaluation, we predicted the open-vocabulary segmentation maps using pre-trained segmentation models and subsequently constructed combined segmentation maps.
% For this process, we leveraged SAN and ODISE when extracting open-vocabulary segmentation maps.
Consequently, for the quantitative evaluation, we use predictions of SAN and ODISE for the open-vocabulary segmentation maps.
Furthermore, the vocabulary set used for the open-vocabulary segmentation models was based on the categories from the COCO-stuff dataset.
Previously, there was no established evaluation protocol capable of accurately measuring the performance of open-vocabulary segmentation and personal visual concepts quantitatively. 
Therefore, we develop combined segmentation maps that enable the concurrent measurement of IoU$^{\text{per}}$ and mIoU.

\begin{figure*}[t!]
    \centering
    \includegraphics[width=0.82\textwidth]{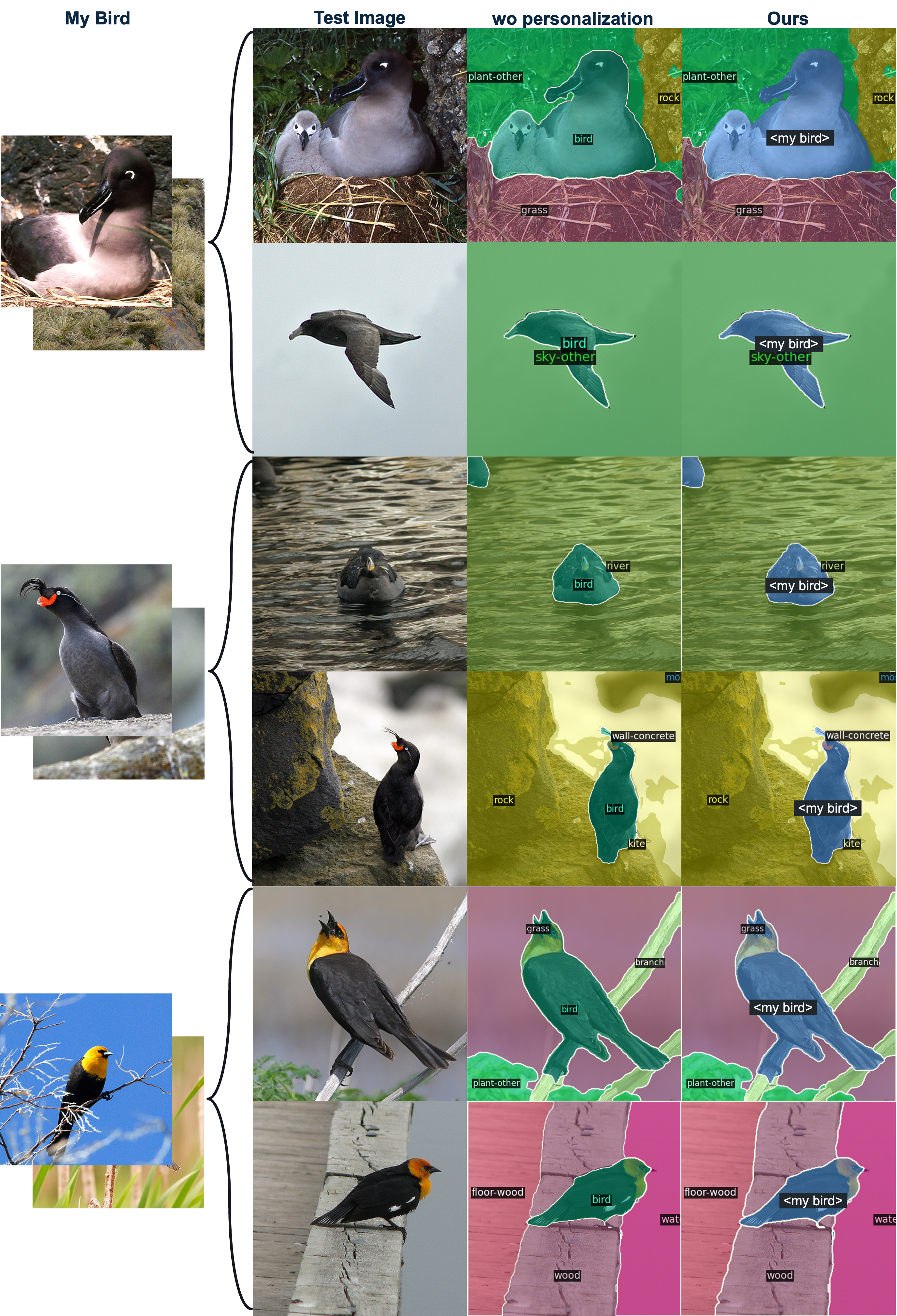}
    \caption{Additional segmentation results on CUB$^\text{per}$. While SAN wihtout personalization fails to capture the personal visual concept (\ie my bird), our method applied to SAN recognizes it.}
    \label{fig:supple_segmentation_cub}
\end{figure*}

% Positive & Negative Images
The FSS-1000 and CUB-200 datasets exhibit high visual similarity within a single category and include foreground masks, making them suitable for assessing personalization performance for specific visual concepts.
The ADE-20K-847 dataset has been extensively used for the quantitative evaluation of existing open-vocabulary segmentation models. 
Thus, we utilize a subset of ADE-20K to construct ADE$^{\text{per}}$ for our experiments.
As described in the main paper, we intentionally include an equal number of images with and without personal visual concepts since models that are fine-tuned to recognize personal concepts overconfidently predict other concepts as personal concepts.

% Category Selection in FSS and ADE
For the CUB$^{\text{per}}$ dataset, all 200 classes are used for evaluation, whereas for the FSS$^{\text{per}}$ and ADE$^{\text{per}}$ datasets, only 30 classes are selected for the experiments, as described in the main paper.
From the FSS-1000 and ADE-20K-847 datasets, we select categories that are challenging to recognize based solely on text descriptions, are rare, or can be easily confused with other categories.
In the FSS dataset, we create pairs of similar categories and use images from these similar yet distinct categories as images without the personal concept (\eg wooden boat$-$banana boat).
Table~\ref{tab: category} describes the categories we selected for FSS$^{\text{per}}$ and ADE$^{\text{per}}$ datasets.

% Selected Classes
\begin{table}[t!]
    \centering
    \begin{tabular}{c|p{0.75\linewidth}<{\centering}}
    \toprule
    Dataset & Categories \\
    \drule
    FSS$^{\text{per}}$ & adidas logo1$-$jordan logo / apple icon$-$yonex icon / banana boat$-$wooden boat / bath ball$-$pokermon ball / bulbul bird$-$chickadee bird / cactus ball$-$gym ball / croquet ball$-$french ball / esport chair$-$ganeva chair / folding chair$-$hair razor / golf ball$-$soccer ball / hair drier$-$rocking chair / jay bird$-$magpie bird / kappa logo$-$nike logo / kobe logo$-$puma logo / ping-pong ball$-$rugby ball  \\
    \midrule
    ADE$^{\text{per}}$ & altarpiece / ashtray / banner / beacon / booklet / candelabrum / canister / chest / console table / crane / dirt track / easel / embankment / footbridge / hot tub / hovel / kettle / kitchen island / pane / parking meter / pier / place mat / postbox / rod / runway / saucepan / shower stall / soap dispenser / stretcher / towel rack \\
    \bottomrule
    \end{tabular}
    \caption{The categories we selected for FSS$^{\text{per}}$ and ADE$^{\text{per}}$.}
    \label{tab: category}
\end{table}

\section{Additional Experiments}

\input{tables/supp-miou}

\subsection{Hyper-parameter Sensitivity}
Fig.~\ref{fig:hyperparams} demonstrates the sensitivity of the hyper-parameters used in our work. 
Our proposed method includes the following hyper-parameters: 1) interpolation value between visual and textual embeddings denoted as $\alpha$, lambda values for $\mathcal{L}^{\text{neg}}_{\text{M}}$ and $\mathcal{L}^{\text{neg}}_{\text{Z}}$ denoted as $\lambda^{\text{neg}}_{\text{M}}$, and $\lambda^{\text{neg}}_{\text{Z}}$, respectively.
For the model and dataset in the experiments, we used SAN and CUB-200, respectively.
Regarding $\alpha$, we empirically found that we achieve promising performances when $\alpha$ is set to low values.
This indicates that while injecting visual information indeed improves performances, the information of textual embedding need to be included more than that of visual embedding. 
We select $\alpha=0.1$ since it achives the best IoU$^{per}$.
Also, the performances saturate as $\lambda^{\text{neg}}_{\text{M}}$ reaches to a certain value, so we select ($\lambda^{\text{neg}}_{\text{M}}=500$) when further performance gain is no longer observed.
Additionally, using different values of $\lambda^{\text{neg}}_{\text{Z}}$ shows consistent performances, so we select the value ($\lambda^{\text{neg}}_{\text{Z}}=0.1$) with the best performance.

\subsection{Segmentation Results}

Fig.~\ref{fig:supple_segmentation_cub} compares the segmentation results of SAN without personalization and our method applied to SAN using CUB$^\text{per}$ as the dataset, which demonstrate the effectiveness of understanding personal visual concept (\ie my bird).
For SAN without personalization, we provide the text descriptions of the visual concept we want to personalize. 
Such a qualitative analysis clearly demonstrates that our newly proposed task, personalized open-vocabulary semantic segmentation, is challenging with the existing open-vocabulary segmentation model, and it needs to be explored.
We believe that our study serves as a cornerstone to further improve performance on personalized open-vocabulary semantic segmentation task.

\subsection{Comparison with Visual Embedding}

Since our work is the first to propose the personalized open-vocabulary semantic segmentation task, we lacked baseline models for comparison. 
In order to further demonstrate the effectiveness of our proposed approach, we compared our method against a baseline that uses masked visual embeddings instead of text embeddings. 
Specifically, given few training images and masks corresponding to a given personal concept, we extracted visual embeddings using the image encoder of CLIP and replaced the text embeddings of personal concept with the averaged visual embeddings.
% In other words, we replaced the text embedding with the averaged visual embeddings while maintaining the other components of open-vocabulary segmentation models (\textit{e.g.}, SAN, ODISE).
During the experiment, we maintained the other components of open-vocabulary segmentation models (\textit{e.g.}, SAN, ODISE).

Table~\ref{supp_tab:IoU} shows that our method significantly outperforms the baseline using masked visual embeddings on the FSS$^\text{per}$ and CUB$^\text{per}$ datasets. 
This result demonstrates the necessity of using the representation space of text encoders for tasks related to open-vocabulary segmentation.
Similar results are observed in the $\alpha$ ablation study in Fig.~\ref{fig:hyperparams}, where the performance in IoU$^\text{per}$ drops dramatically as the proportion of masked visual embeddings increases excessively.

\input{tables/supp-concat}

\subsection{Further Experiments on Distinguishing \\ Personal Concept from Similar Classes}

\begin{figure*}[t!]
    \centering
    \includegraphics[width=1.0\linewidth]{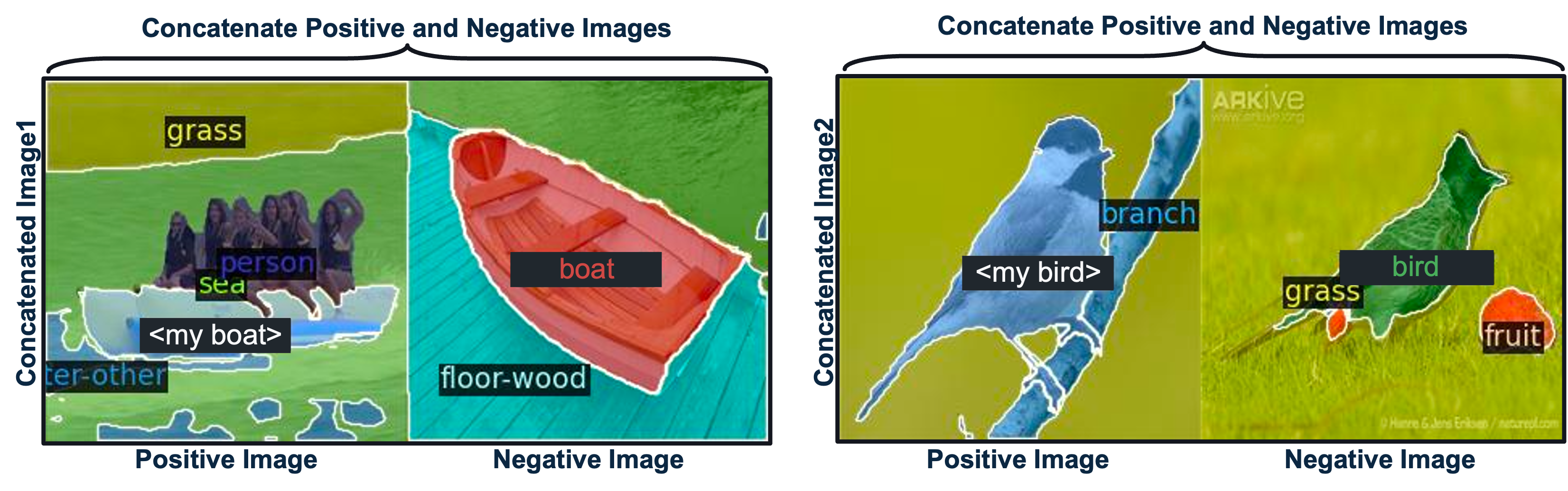}
    \caption{Test images of \textbf{concat} dataset (FSS$^{per}$). These datasets are used for evaluating the performance on distinguishing between the target visual concept (\textit{e.g.}, ``my boat'', ``my bird'') and its corresponding similar classes (\textit{e.g.}, ``boat'', ``bird'') within the same image.}
    \label{fig:supple_concat}
\end{figure*}

As shown in the qualitative results of the main paper (Fig.~3, 5, and 6), our method can distinguish between the target visual concept (\textit{e.g.}, ``my boat'', ``my bird'') and its corresponding similar classes (\textit{e.g.}, ``boat'', ``bird'') within the same image. 
For the results, we used FSS and CUB datasets annotated with fine-grained segmentation maps, which include several images and masks on specific fine-grained classes. 
However, we found that these datasets often contain only one object per image, which limits our evaluation on distinguishing the personal concept with its corresponding similar classes within the same image. 

To this end, we conducted an additional experiment by concatenating two images horizontally: 1) a positive image that contains the target visual concept and 2) a negative image that contains the similar class but without the target visual concept. 
Then, we treated the concatenated images as a single image and input it for segmentation.
For better understanding, we show examples of concatenated positive and negative class images, which we refer to as the ``concat dataset'' in Fig.~\ref{fig:supple_concat}.
Table~\ref{supp_tab:concat} demonstrates that our approach significantly improves IoU$^\text{per}$ compared to existing open-vocabulary segmentation models in such an experimental setting. 
These results show that our method can effectively distinguish between the target visual concept and similar but different classes, even when negative classes are present in the same image.

While we used horizontal concatenation to include the negative image in a single image, we believe that constructing datasets with images that naturally include both the target visual concept and its similar classes would further enhance the evaluation of personalized open-vocabulary semantic segmentation.

\end{document}

%% file: tables/main-iou.tex
\begin{table*}[t]
\centering
\begin{center}
{\resizebox{0.75\textwidth}{!}{
{
\begin{tabular}{c||l|ccc|c|ccc|c} 
\toprule
\multirow{2}{*}{Dataset} & \multirow{2}{*}{Method} & \multicolumn{4}{c|}{IoU$^{\text{per}}$} & \multicolumn{4}{c}{mIoU} \\
                         &                         & $K=1$ & $K=3$ & $K=5$ & Avg. & $K=1$ & $K=3$ & $K=5$ & Avg.\\
\drule

& ODISE~\cite{odise} & 10.69 & 10.69 & 10.69 & 10.69
                     & \textbf{23.86} & \textbf{23.86} & \textbf{23.86} & \textbf{23.86} \\
\rowcolor{gray!10} \cellcolor{white}
& + Ours             & \textbf{30.97} & \textbf{33.11} & \textbf{34.05} & \textbf{32.71}
                     & 21.83 & 22.68 & 22.94 & 22.48 \\
\cmidrule{2-10}
& SAN~\cite{san}     & 41.08 & 41.08 & 41.08 & 41.08 
                     & 55.68 & 55.68 & 55.68 & 55.68 \\
\rowcolor{gray!10} \multirow{-4}{*}{\cellcolor{white} FSS$^{\text{per}}$} 
& + Ours             & \textbf{49.80} & \textbf{54.09} & \textbf{56.80} & \textbf{53.56}
                     & \textbf{56.40} & \textbf{56.73} & \textbf{55.85} & \textbf{56.32} \\

\midrule

& ODISE~\cite{odise} & 0.02 & 0.02 & 0.02 & 0.02 
                     & \textbf{47.48} & \textbf{47.48} & \textbf{47.48} & \textbf{47.48} \\
\rowcolor{gray!10} \cellcolor{white}
& + Ours             &\textbf{5.39} &\textbf{5.90} &\textbf{5.99} &\textbf{5.76}
                     & 45.16 & 44.94 & 44.88 & 44.99 \\
\cmidrule{2-10} \cellcolor{white}
& SAN~\cite{san}     & 68.25 & 68.25 & 68.25 & 68.25
                     & 77.32 & 77.32 & 77.32 & 77.32 \\
\rowcolor{gray!10} \multirow{-4}{*}{\cellcolor{white} CUB$^{\text{per}}$} 
& + Ours             & \textbf{76.70} &\textbf{77.21} &\textbf{76.80} &\textbf{76.90} 
                     & \textbf{77.36} &\textbf{77.85} &\textbf{78.29} &\textbf{77.83} \\

\midrule

& ODISE~\cite{odise} & 1.24 & 1.24 & 1.24 & 1.24
                     & \textbf{12.22} & \textbf{12.22} & \textbf{12.22} & \textbf{12.22} \\
\rowcolor{gray!10} \cellcolor{white}
& + Ours             & \textbf{9.45} & \textbf{10.85} & \textbf{13.43} & \textbf{11.24}
                     & 12.21 & 12.18 & 12.18 & 12.19 \\
\cmidrule{2-10}
& SAN~\cite{san}     & 6.88 & 6.88 & 6.88 & 6.88
                     & 17.20 & 17.20 & \textbf{17.20} & 17.20 \\
\rowcolor{gray!10} \multirow{-4}{*}{\cellcolor{white} ADE$^{\text{per}}$} 
& + Ours             & \textbf{17.08} & \textbf{24.80} & \textbf{26.15} & \textbf{22.67} 
                     & \textbf{17.26} & \textbf{17.21} & 17.19 & \textbf{17.22} \\

\bottomrule

\end{tabular}}}}
\end{center}
\vspace{-0.6cm}
\caption{Comparisons on IoU$^{\text{per}}$ and mIoU. We apply our method on both SAN~\cite{san} and ODISE~\cite{odise} on FSS$^{\text{per}}$, CUB$^{\text{per}}$, and ADE$^{\text{per}}$. We vary $K$, the number of images and masks, to 1, 3, and 5.}
\vspace{-0.4cm}
\label{tab:IoU}
\end{table*}

%% file: tables/main-ablation.tex
\begin{table}[t!]
\centering

\begin{center}
\resizebox{0.47\textwidth}{!}{
\begin{tabular}{p{0.05\textwidth}<{\centering}p{0.05\textwidth}<{\centering}p{0.05\textwidth}<{\centering}|p{0.07\textwidth}<{\centering}p{0.07\textwidth}<{\centering}p{0.07\textwidth}<{\centering}p{0.07\textwidth}<{\centering}} 

\toprule
Text Prompt & Neg. Mask & Visual Inject & \multirow{2}{*}{mIoU} & \multirow{2}{*}{IoU$^{\text{per}}$} & \multirow{2}{*}{IoU$^{\text{per}}_{\text{precision}}$} & \multirow{2}{*}{IoU$^{\text{per}}_{\text{recall}}$} \\
\drule
- & - & -                            & 77.32 & 68.25 & 92.25 & 72.95 \\
\checkmark & - & -                   & 77.89 & 69.70 & 74.75 & 91.04 \\ 
\checkmark & \checkmark & -          & 77.89 & 73.71 & 80.07 & 90.17 \\ 
\checkmark & - & \checkmark          & 77.65 & 65.94 & 70.06 & 91.58 \\ 
\checkmark & \checkmark & \checkmark & \textbf{78.29} & \textbf{76.80} & 84.51 & 89.07 \\ 
\bottomrule
\end{tabular}}
\end{center}
\vspace{-0.5cm}
\caption{Ablation study on our proposed method. We show how each module contributes to the performance gain by reporting mIoU, IoU$^{\text{per}}$, IoU$^{\text{per}}_{\text{precision}}$ and IoU$^{\text{per}}_{\text{recall}}$.}
\vspace{-0.4cm}
\label{tab:ablation}
\end{table}

%% file: tables/main-knn_clip.tex
\begin{table}[h!]
\vspace{-0.3cm}
\centering
{\resizebox{0.35\textwidth}{!}{
{
\begin{tabular}{c||l|ccc} 
\toprule
Dataset & Metric  & IoU$^{\text{per}}$ & mIoU\\
\drule
\multirow{2}{*}{\makecell{CUB$^{\text{per}}$ \\ ($K=5$)}} 
& kNN-CLIP      & 1.27 & 56.34 \\
& Ours          & \textbf{76.80} & \textbf{78.29} \\
\bottomrule
\end{tabular}}}}
\vspace{-0.3cm}
\caption{Comparisons with kNN-CLIP on CUB$^{\text{per}}$ with $K=5$.}
\label{tab:knn_clip}
\end{table}

%% file: tables/main-few_shot.tex
\begin{table}[h!]
\vspace{-0.6cm}
\centering
{\resizebox{0.4\textwidth}{!}{
{
\begin{tabular}{c||l|ccc} 
\toprule
Dataset & Methods & IoU$^{\text{per}}$ & IoU$^{\text{per}}_{\text{precision}}$ & IoU$^{\text{per}}_{\text{recall}}$ \\
\drule
\multirow{3}{*}{\makecell{CUB$^{\text{per}}$ \\ ($K=1$)}} 
& SEEM   & 45.54 & 47.64 & 90.82 \\
& SegGPT & 47.00 & 48.56 & \textbf{93.88} \\
& Ours   & \textbf{76.80} & \textbf{84.51} & 89.07 \\
\bottomrule
\end{tabular}}}}
\vspace{-0.3cm}
\caption{Comparisons with recent few-shot semantic segmentation methods on CUB$^{\text{per}}$ with $K=1$.}
\label{tab:few_shot}
\end{table}

%% file: tables/supp-miou.tex
\begin{table*}[t]
\centering
\begin{center}
{\resizebox{0.9\textwidth}{!}{
{
\begin{tabular}{c||l|ccc|c|ccc|c} 
\toprule
\multirow{2}{*}{Dataset} & \multirow{2}{*}{Method} & \multicolumn{4}{c|}{IoU$^{\text{per}}$} & \multicolumn{4}{c}{mIoU} \\
                         &                         & $K=1$ & $K=3$ & $K=5$ & Avg. & $K=1$ & $K=3$ & $K=5$ & Avg.\\
\drule

& ODISE~\cite{odise} & 10.69 & 10.69 & 10.69 & 10.69
                     & \textbf{23.86} & \textbf{23.86} & \textbf{23.86} & \textbf{23.86} \\

& + Visual Embedding & 0.00 & 0.00 & 0.00 & 0.00
                     & 23.37 & 23.37 & 23.37 & 23.37 \\
                     
\rowcolor{gray!10} \cellcolor{white}
& + Ours             & \textbf{30.97} & \textbf{33.11} & \textbf{34.05} & \textbf{32.71}
                     & 21.83 & 22.68 & 22.94 & 22.48 \\
\cmidrule{2-10}
& SAN~\cite{san}     & 41.08 & 41.08 & 41.08 & 41.08 
                     & 55.68 & 55.68 & 55.68 & 55.68 \\

& + Visual Embedding & 0.00 & 0.00 & 0.00 & 0.00
                     & 56.12 & 55.36 & 54.45 & 55.31 \\

\rowcolor{gray!10} \multirow{-6}{*}{\cellcolor{white} FSS$^{\text{per}}$} 
& + Ours             & \textbf{49.80} & \textbf{54.09} & \textbf{56.80} & \textbf{53.56}
                     & \textbf{56.40} & \textbf{56.73} & \textbf{55.85} & \textbf{56.32} \\

\midrule

& ODISE~\cite{odise} & 0.02 & 0.02 & 0.02 & 0.02 
                     & \textbf{47.48} & \textbf{47.48} & \textbf{47.48} & \textbf{47.48} \\

& + Visual Embedding & 0.02 & 0.02 & 0.02 & 0.02
                     & 47.71 & 47.36 & 47.31 & 47.46 \\

\rowcolor{gray!10} \cellcolor{white}
& + Ours             &\textbf{5.39} &\textbf{5.90} &\textbf{5.99} &\textbf{5.76}
                     & 45.16 & 44.94 & 44.88 & 44.99 \\
\cmidrule{2-10} \cellcolor{white}
& SAN~\cite{san}     & 68.25 & 68.25 & 68.25 & 68.25
                     & 77.32 & 77.32 & 77.32 & 77.32 \\

& + Visual Embedding & 0.70 & 0.00 & 0.00 & 0.23
                     & 72.77 & 67.13 & 65.98 & 68.63 \\

\rowcolor{gray!10} \multirow{-6}{*}{\cellcolor{white} CUB$^{\text{per}}$} 
& + Ours             & \textbf{76.70} &\textbf{77.21} &\textbf{76.80} &\textbf{76.90} 
                     & \textbf{77.36} &\textbf{77.85} &\textbf{78.29} &\textbf{77.83} \\

% \midrule

% & ODISE~\cite{odise} & 1.24 & 1.24 & 1.24 & 1.24
%                      & \textbf{12.22} & \textbf{12.22} & \textbf{12.22} & \textbf{12.22} \\
% \rowcolor{gray!10} \cellcolor{white}
% & + Ours             & \textbf{9.45} & \textbf{10.85} & \textbf{13.43} & \textbf{11.24}
%                      & 12.21 & 12.18 & 12.18 & 12.19 \\
% \cmidrule{2-10}
% & SAN~\cite{san}     & 6.88 & 6.88 & 6.88 & 6.88
%                      & 17.20 & 17.20 & \textbf{17.20} & 17.20 \\
% \rowcolor{gray!10} \multirow{-4}{*}{\cellcolor{white} ADE$^{\text{per}}$} 
% & + Ours             & \textbf{17.08} & \textbf{24.80} & \textbf{26.15} & \textbf{22.67} 
%                      & \textbf{17.26} & \textbf{17.21} & 17.19 & \textbf{17.22} \\

\bottomrule

\end{tabular}}}}
\end{center}
\vspace{-0.2cm}
\caption{Comparisons with baseline using visual embedding. We apply our method on both SAN~\cite{san} and ODISE~\cite{odise} on FSS$^{\text{per}}$ and CUB$^{\text{per}}$. We vary $K$, the number of images and masks, to 1, 3, and 5.}
\vspace{-0.2cm}
\label{supp_tab:IoU}
\end{table*}

%% file: tables/supp-concat.tex
\begin{table}[t]
\centering
\begin{center}
\begin{tabular}{c||l|cc} 
\toprule
Dataset & Method & IoU$^{\text{per}}$ & mIoU \\
\drule

& SAN~\cite{san}     & 28.59 & 57.66 \\
\rowcolor{gray!10} \multirow{-2}{*}{\cellcolor{white} FSS$^{\text{per}}$} 
& + Ours             & \textbf{38.47} & \textbf{60.09} \\

\midrule

\cellcolor{white}
& SAN~\cite{san}     & 24.25 & 81.62 \\
\rowcolor{gray!10} \multirow{-2}{*}{\cellcolor{white} CUB$^{\text{per}}$} 
& + Ours             & \textbf{40.63} &\textbf{81.71} \\

\bottomrule
\end{tabular}
\end{center}
\vspace{-0.2cm}
\caption{Quantative results on \textbf{concat} datasets. We apply our method on SAN~\cite{san} on FSS$^{\text{per}}$ and CUB$^{\text{per}}$. $K$ is set to 5.}
\vspace{-0.2cm}
\label{supp_tab:concat}
\end{table}